%% file: main.tex
\documentclass[runningheads]{llncs}
\usepackage[T1]{fontenc}
\usepackage{graphicx,verbatim}
\usepackage{booktabs}
\usepackage[table]{xcolor}
\usepackage{pgfplots}
\usepackage{multirow}

\pgfplotsset{compat=1.18}
\usepackage{tikz,amsmath,amssymb}
\usetikzlibrary{arrows.meta,positioning,calc,fit,backgrounds,decorations.pathreplacing}

\usetikzlibrary{
    arrows.meta,     % >=Stealth
    positioning,     % right=of, below=of
    calc,            % coordinate calculations like -|
    fit,             % fit=() for highlight box
    backgrounds      % on background layer
}
\usetikzlibrary{arrows.meta, positioning, calc, fit, backgrounds}

\usepackage{amssymb}
\usepackage{textcomp}
\usepackage[table]{xcolor}
\usepackage{hyperref} 
\usepackage{orcidlink}
\hypersetup{
    colorlinks,
    linkcolor={red!50!black},
    citecolor={blue!50!black},
    urlcolor={blue!50!black}
}

\begin{document}
\title{%Toward Agentic Dataset Curation in Histopathology: Vision-Language Models for 
Vision-Language Models as Zero-Annotation Oracles  in Histopathology}
\titlerunning{Zero-Annotation Oracles in Histopathology}
\author{Vishal Jain\inst{1}\orcidlink{0009-0001-3447-9211} \and
Giorgio Buzzanca\inst{2}\orcidlink{0009-0009-0340-8822} \and
Sarah Cechnicka\inst{1}\orcidlink{0009-0008-3449-9379} \and
Maarten Naesens\inst{3}\orcidlink{0000-0002-5625-0792} \and
Priyanka Koshy\inst{4}\orcidlink{0000-0002-2313-5122} \and
Tri Nguyen\inst{5}\orcidlink{0000-0001-6475-0706} \and
Jesper Kers\inst{2}\orcidlink{0000-0002-2418-5279} \and
Candice Roufosse\inst{1}\orcidlink{0000-0002-6490-4290} \and
Bernhard Kainz\inst{1,6}\orcidlink{0000-0002-7813-5023}}
\authorrunning{V. Jain et al.}
\institute{Imperial College London, London, UK\\
    \email{v.jain24@imperial.ac.uk}
    \and Leiden University Medical Center, Leiden, The Netherlands
    \and KU Leuven, Leuven, Belgium
    \and University Hospitals Leuven, Leuven, Belgium
    \and University Medical Center Utrecht, Utrecht, The Netherlands
    \and Dept. AIBE, Friedrich-Alexander University Erlangen-N\"urnberg, Germany}

\maketitle              % typeset the header of the contribution
\begin{abstract}
Foreground segmentation is the critical first step of every computational pathology pipeline, yet existing methods rely on hand-tuned heuristics or supervised models that overfit to narrow stain and scanner distributions, failing silently on specialised stains such as Jones silver or Elastica van Gieson. We propose a coarse-to-fine approach that recasts foreground segmentation as a visual perception task and leverages general-purpose vision-language models (VLMs) as zero-annotation oracles. Our key insight is that tissue-versus-background discrimination is a natural-image recognition problem, not a histopathological one, so VLMs trained on internet-scale corpora generalise where domain-specific models cannot. We introduce \textit{Leica-75}, a benchmark of 75 renal transplant whole-slide images spanning three stain families. On Leica-75, our method achieves the highest segmentation quality on out-of-distribution stains (Dice $0.858_{\pm 0.027}$ on Jones, $0.853_{\pm 0.041}$ on EVG) with $7\times$ lower cross-stain variance than the best supervised baseline, while remaining competitive on in-distribution H\&E. Few-shot prompting with automatically curated exemplars (\textit{Auto-context}) rescues hard cases on \textit{Stress-32} ($n{=}32$), a curated stress-test subset (Dice $0.470 \rightarrow 0.819$ for the 2B model). VLM-based annotation review matches human expert consensus ($\kappa{=}0.989$ for blur detection; mean precision/recall grading accuracy $0.708$ vs.\ human $0.646$ for segmentation mask review). The resulting pseudo-labels are used to distil lightweight student models that are as performant as the teacher model while running for a fraction of the cost. Our framework provides a principled, scalable solution to a persistent infrastructure bottleneck in digital pathology.
Code is available at \url{https://github.com/VishalJ99/vlm-wsi-auto-context}

% Authors must provide keywords and are not allowed to remove this Keyword section.

\end{abstract}
%
%
%\section{Introduction}
%\label{sec:intro}

\section{Introduction}
\label{sec:intro}

Every computational pathology pipeline begins by separating tissue from background in a whole-slide image (WSI). At gigapixel scale, downstream methods typically operate on fixed-size patches and aggregate representations via multiple-instance learning (MIL)~\cite{lu2021data,campanella2019clinical,ilse2018attention}, creating an extreme class-imbalance setting in which diagnostically relevant regions may occupy only a small fraction of the slide~\cite{bejnordi2017diagnostic}. Including background or artefacts among candidate patches can induce spurious correlations in attention mechanisms and degrade performance~\cite{howard2021impact,ye2024cleverhans}. Reliable foreground segmentation is therefore a foundational prerequisite whose errors propagate to all subsequent analysis.

In practice, tissue preparation and digitisation introduce substantial variation across institutions, scanners, and chemical protocols~\cite{howard2021impact,janowczyk2019histoqc,dejong2025currentpathologyfoundationmodels}. Classical methods based on fixed thresholds or unsupervised clustering (Otsu~\cite{otsu1979threshold}, $K$-means~\cite{MacQueen1967SomeMF}) are brittle under distribution shift. Rule-based pipelines such as HistoQC~\cite{janowczyk2019histoqc} require manual recalibration per site, while supervised models such as HEST~\cite{jaume2024hest} and GrandQC~\cite{weng2024grandqc} can overfit to their training distribution. A major contributor is limited stain diversity in widely used pathology tooling and current foundation models~\cite{chen2024uni,vorontsov2024virchow,lu2024visual,nechaev2024hibou,filiot2024phikon}, which are typically trained on H\&E-dominated datasets. This induces stain bias that is consequential in clinical workflows employing specialised stains such as Jones methenamine silver, Elastica van Gieson (EVG), and periodic acid-Schiff (PAS)~\cite{howard2021impact,dejong2025currentpathologyfoundationmodels,lin2025diagnosticperformancerevealingquantifying}. %On our benchmark, GrandQC achieves Dice $0.870_{\pm0.047}$ on H\&E but drops to $0.775_{\pm0.194}$ on Jones; HEST is more stable but plateaus lower ($0.807_{\pm0.075}$ on H\&E, $0.743_{\pm0.109}$ on Jones), and HistoQC can collapse entirely (Dice $0.224_{\pm0.235}$ on H\&E, $0.281_{\pm0.298}$ on Jones).

For the purpose of WSI preprocessing, tissue-versus-background discrimination depends primarily on generic visual cues (colour, texture, and structural regularity) rather than histopathological semantics. This suggests that general-purpose vision-language models (VLMs)~\cite{comanici2025gemini25pushingfrontier,bai2025qwen3vltechnicalreport}, trained on internet-scale corpora, can provide a strong prior for stain- and site-agnostic foreground detection. If such a prior can be leveraged reliably, it offers a path to cross-site deployment without the per-institution tuning that H\&E-centric pipelines often require. The remaining obstacle is efficiency: na\"ively applying VLM inference at WSI scale is prohibitively expensive. We therefore introduce a coarse-to-fine method that uses VLMs as zero-annotation oracles while aggressively restricting the number of expensive calls, and we use the resulting pseudo-labels to distil lightweight student models for fast, low-cost deployment.

Therefore, our \textbf{contributions} are:
\textbf{(1)} A coarse-to-fine method that leverages general-purpose VLMs as zero-annotation oracles for WSI foreground segmentation, and a distillation pathway that converts VLM pseudo-labels into lightweight student models suitable for deployment.
\textbf{(2)} The first systematic evaluation of general-purpose VLMs as zero-annotation oracles for stain-agnostic foreground segmentation across three stain families, showing improved robustness on out-of-distribution stains and demonstrating that automated few-shot prompting can rescue hard cases while narrowing the gap between smaller and larger VLMs.
\textbf{(3)} Evidence that VLMs can provide practical quality control at scale on two tasks: (i) near-perfect binary out-of-focus patch triage, and (ii) precision/recall mask review on a synthetic corruption benchmark, where VLM reviewers match or exceed the strongest human rater and show competitive agreement with expert cohorts.

\noindent\textbf{Related work.}
Foreground segmentation has progressed from classical thresholding and clustering~\cite{otsu1979threshold,MacQueen1967SomeMF} and rule-based quality control (HistoQC~\cite{janowczyk2019histoqc}) to supervised pipelines embedded in computational pathology toolkits such as TIAToolbox~\cite{pocock2022tiatoolbox} and TRIDENT~\cite{zhang2025accelerating}, which incorporate GrandQC~\cite{weng2024grandqc} and HEST~\cite{jaume2024hest}. However, these models are typically trained predominantly on H\&E and inherit stain-specific biases that can degrade performance on specialised stains~\cite{dejong2025currentpathologyfoundationmodels,howard2021impact,ye2024cleverhans}. VLMs remain largely unexplored for histopathology preprocessing; the closest work is CORE~\cite{nasir_core_2025}, which uses Florence-2~\cite{xiao_florence-2_2023} with SAM~\cite{kirillov_segment_2023} for tissue mask extraction for WSI registration but falls back to a supervised U-Net when staining quality degrades. In contrast, we treat  segmentation as a general visual perception task solvable via text prompting of general-purpose VLMs~\cite{bai2024qwen2vl,bai2025qwen25vl}, extend this to automated exemplar selection for few-shot refinement, and use VLM-generated pseudo-labels to distil lightweight students. We emphasize patch-level tissue labels, aligning the segmentation output with the primary consumer in modern pipelines: MIL-based WSI aggregation~\cite{ilse2018attention,campanella2019clinical,shao2021transmil}, which typically requires patch-level filtering rather than pixel-accurate boundaries.

\section{Method}
\label{sec:method}

%We frame foreground segmentation as a coarse-to-fine inference problem in which a general-purpose VLM serves as an annotation oracle whose outputs are progressively refined and ultimately distilled into a lightweight deployment model. 
Our method (Fig.~\ref{fig:pipeline}) is a coarse-to-fine approach that produces patch-level tissue labels using a general-purpose VLM as a zero-annotation oracle, and then distills these pseudo-labels into a lightweight deployment model. The core method comprises seven stages:
(i) thumbnail-level VLM box grounding,
(ii) two-pass colour clustering to obtain a coarse candidate foreground mask,
(iii) zero-shot patch classification (tissue/background), jointly producing focus-quality labels as a side-output,
(iv) \textit{Auto-context}: VLM-guided sampling of candidate points and automatic selection of high-magnification tissue/background exemplars,
(v) few-shot patch re-classification using the exemplars, (vi) morphological mask refinement
and (vii) hard-label distillation into a student model.
Separately, we use an auxiliary VLM reviewer to grade mask precision/recall and to trigger stages (iv-v) when refinement is required.
\input{figures/overview2}
% ----------------------------------------------------------------------------
% Stage 1: VLM-guided tissue core localisation
% ----------------------------------------------------------------------------

\noindent\textbf{VLM-guided tissue core localisation.}
Given a WSI $\mathbf{W}$ at native resolution $H_0 \times W_0$, we extract a
thumbnail $\mathbf{T} \in \mathbb{R}^{h \times w \times 3}$ at a fixed
downsampling factor $d$ (typically $d{=}64$, yielding $h{=}H_0/d$,
$w{=}W_0/d$). Rather than relying on hand-crafted colour thresholds, we
query a VLM with $\mathbf{T}$ and a structured prompt requesting bounding
boxes around tissue cores visible in the thumbnail. The VLM returns a set
of axis-aligned bounding boxes
$\mathcal{B} = \{b_i\}_{i=1}^{B}$, where each
$b_i = (x_i^{\min}, y_i^{\min}, x_i^{\max}, y_i^{\max})$ specifies the
top-left and bottom-right corners of a tissue core in thumbnail
coordinates. These bounding boxes define the regions of interest for
the subsequent colour-based segmentation.%, restricting downstreamprocessing to slide regions that contain diagnostically relevant material.

% ----------------------------------------------------------------------------
% Stage 2: Two-pass colour segmentation
% ----------------------------------------------------------------------------
\noindent\textbf{Two-pass colour segmentation.}
We refine the thumbnail boxes from Stage~1 using $K$-means in CIELAB space.

\noindent\emph{Global pass.}\;
We cluster all thumbnail pixels into $K{=}2$ clusters with assignments $z_{ij}\in\{1,2\}$ and centroids $\{\boldsymbol{\mu}_1,\boldsymbol{\mu}_2\}$. We identify the global background centroid as the majority cluster,
\begin{equation}
k_{\mathrm{bg}}=\arg\max_k \big|\{(i,j): z_{ij}=k\}\big|,  \qquad
\boldsymbol{\mu}_{\mathrm{bg}}=\boldsymbol{\mu}_{k_{\mathrm{bg}}}.
\end{equation}

\noindent\emph{Local pass.}\;
For each VLM box $b_i\in\mathcal{B}$ we cluster pixels in the crop $\mathbf{T}_i$ into $K{=}3$ clusters with centroids $\{\boldsymbol{\mu}^{(i)}_k\}_{k=1}^3$. The crop background cluster is chosen by nearest-centroid matching,
\begin{equation}
%$
k_{\mathrm{bg}}^{(i)}=\arg\min_k \left\|\boldsymbol{\mu}^{(i)}_k-\boldsymbol{\mu}_{\mathrm{bg}}\right\|_2 ,
%$
\label{eq:local-bg}
\end{equation}
and the binary crop mask is $\mathbf{M}^{(i)}=\mathbf{1}[z^{(i)}\neq k_{\mathrm{bg}}^{(i)}]$. Using $K{=}3$ locally reduces failures when a dominant artefact would otherwise absorb a cluster under $K{=}2$.

\noindent We combine crop masks via 
%\[
$
\mathbf{M}^{(1)}(u,v)=\max_{i\in\{1,\dots,B\}} \mathbf{M}^{(i)}(u,v),
$
%\]
and upsample $\mathbf{M}^{(1)}$ to level-0 coordinates to restrict subsequent VLM evaluation.

% ----------------------------------------------------------------------------
% Stage 3: VLM-guided patch classification
% ----------------------------------------------------------------------------
\noindent\textbf{Zero-shot patch classification.}
We partition the tissue regions identified by $\mathbf{M}^{(1)}$ into a
regular grid of non-overlapping patches $\{x_n\}_{n=1}^{N}$ of size
$512 \times 512$ pixels at level-0. Each patch is presented to the VLM with
a structured prompt requesting two labels: (\emph{i})~a binary tissue
classification $\hat{y}_n^{(0)} \in \{0,1\}$, and (\emph{ii})~a
focus-quality assessment $q_n \in \{\textit{sharp},\, \textit{mild blur},\,
\textit{out of focus}\}$. Since both labels are extracted from a single VLM
call, the blur map is obtained at no additional inference cost. Aggregating
$\{\hat{y}_n^{(0)}\}$ yields an initial tissue map over the slide, while
aggregating $\{q_n\}$ yields a patch-level focus-quality map.

% ----------------------------------------------------------------------------
% VLM quality review and few-shot gating
% ----------------------------------------------------------------------------
% ----------------------------------------------------------------------------
% Point grounding with TTA for exemplar mining
% ----------------------------------------------------------------------------
\noindent\textbf{VLM-based segmentation mask review.}
Once the tissue mask has been produced, we employ a frontier VLM
(Gemini~\cite{gemini15report,gemini3release}) as an automated annotation
reviewer. The reviewer receives two inputs: (\emph{i})~the tissue-core crop from the WSI thumbnail, and (\emph{ii})~the same crop with the predicted mask overlaid. It is prompted to assess the overall quality of the
segmentation and returns discrete quality grades for both precision and
recall, $g_p, g_r \in \{1,2,3,4\}$. If either grade falls below a
threshold $\tau_g$, our method re-enters the few-shot refinement stage
described in Stages (iv--v). The reviewer thus serves as a scalable quality triage layer, with human
escalation reserved for ambiguous cases.
% ----------------------------------------------------------------------------
% High-magnification exemplar selection
% ----------------------------------------------------------------------------

\noindent\textbf{High-magnification exemplar selection (\textit{Auto-context}).}
Should refinement be triggered (e.g.\ by reviewer grades falling below $\tau_g$),
we query the VLM to perform point grounding on the tissue-core crop, asking it to
identify candidate tissue and background locations. This yields spatially informed
candidate pools $\mathcal{G}^{+}$ (tissue) and $\mathcal{G}^{-}$ (background) that
are better balanced than na\"ive random sampling, which would be dominated by
background. Because point grounding is imprecise, we present the corresponding
high-magnification patches to the VLM in a comparative re-ranking prompt and select
the top-$K$ tissue exemplars $\mathcal{E}^{+}$ and top-$K$ background exemplars
$\mathcal{E}^{-}$, forming the in-context exemplar set
$\mathcal{E}=\mathcal{E}^{+}\cup\mathcal{E}^{-}$.

% ----------------------------------------------------------------------------
% Few-shot classification
% ----------------------------------------------------------------------------
\noindent\textbf{Few-shot patch classification.}
The exemplar set $\mathcal{E}$ is prepended to the VLM prompt as labelled
demonstrations, and all patches are re-classified. The updated label for
patch $n$ is $\hat{y}_n = f_{\text{VLM}}(x_n \mid \mathcal{E})$, where
$f_{\text{VLM}}$ denotes the VLM response under the few-shot prompt. This stage targets whole slide images where zero-shot performance is limited by background staining artefacts whose high-magnification appearance mimics cellular structures.
% ----------------------------------------------------------------------------
% Post processing
% ----------------------------------------------------------------------------

\noindent\textbf{Post-processing.}
%Let $\hat{\mathbf{Y}}\in\{0,1\}^{N_y\times N_x}$ denote the patch-level tissue mask on the $512{\times}512$ grid. 
We apply lightweight morphological cleanup on this grid using 4-neighbour connectivity: (i) remove foreground connected components with area $<8$ patches, and (ii) fill background holes with area $<10$ patches. %This step reduces isolated false positives and small spurious gaps without affecting large tissue regions.
% ----------------------------------------------------------------------------
% Stage 4: Knowledge distillation
% ----------------------------------------------------------------------------

\noindent\textbf{Knowledge distillation.}
We perform \emph{hard-label} distillation from VLM pseudo-labels at patch level. A lightweight
student $f_\theta$ (MobileNetV3-Large-100, 4.2\,M params, 2-class head) is trained with standard
cross-entropy on canonical binary targets $\hat y_n\in\{0,1\}$; no temperature scaling or
teacher-logit matching is used.

\input{evaluation}

\section{Conclusion}
\label{sec:conclusion}

By reframing foreground segmentation as a perception task, we show that general-purpose VLMs outperform dedicated pathology tools on out-of-distribution stains (Dice $0.858$ Jones, $0.853$ EVG) with $7{\times}$ lower cross-stain variance, while few-shot prompting rescues hard cases ($+0.349$ Dice) and VLM reviewers match human consensus ($\kappa{=}0.989$ blur; joint accuracy $0.708$ vs.\ $0.646$). Distillation into lightweight students decouples deployment from frontier-model cost. Our work establishes VLMs as practical zero-annotation oracles for stain-agnostic dataset curation at institutional scale.
%\newpage

\noindent\textbf{Acknowledgements.}
G. Buzzanca acknowledges Leiden University Medical Center. J. Kers acknowledges Leiden University Medical Center and Amsterdam UMC. M. Naesens acknowledges KU Leuven. T. Nguyen acknowledges University Medical Center Utrecht. This project has received funding from the European Union's Horizon Europe research and innovation programme under grant agreement No. 101072891.
V. Jain and S. Cechnicka are supported by the UKRI Centre for Doctoral Training AI4Health (EP/S023283/1). This work was further supported by the European Research Council (ERC) project MIA-NORMAL (101083647), ERC project CHARMS (101246053), the German Research Foundation (DFG, projects 512819079 and 513220538), and the State of Bavaria through HTA and the Bavarian Foundation Model Initiative. HPC resources were provided by NHR@FAU of FAU Erlangen-N\"urnberg under the NHR projects b143dc and b180dc. NHR funding is provided by federal and Bavarian state authorities, and NHR@FAU hardware is partially funded by the DFG under project 440719683. 
The authors acknowledge the use of resources provided by the Isambard-AI National AI Research Resource (AIRR). Isambard-AI is operated by the University of Bristol and is funded by the UK Government's Department for Science, Innovation and Technology (DSIT) via UK Research and Innovation, and the Science and Technology Facilities Council [ST/AIRR/I-A-I/1023]~\cite{mcintoshsmith2024isambardai}. 
Dr. Roufosse is supported by the National Institute for Health Research (NIHR) Biomedical Research Centre based at Imperial College Healthcare NHS Trust and Imperial College London (ICL). The views expressed are those of the authors and not necessarily those of the NHS, the NIHR, or the Department of Health. Dr. Roufosse's research activity is made possible with generous support from Sidharth and Indira Burman. 
Human samples used in this research project were obtained from the Imperial College Healthcare Tissue \& Biobank (ICHTB). ICHTB is supported by the NIHR Biomedical Research Centre based at Imperial College Healthcare NHS Trust and ICL. ICHTB is approved by Wales REC3 to release human material for research (22/WA/2836).

\subsubsection{\discintname}
The authors declare no competing interests relevant to the content of this article.

\bibliographystyle{splncs04}
\bibliography{references}

\end{document}

%% file: figures/overview2.tex
% TikZ pipeline figure
\begin{figure}[t]
\centering
\resizebox{\textwidth}{!}{%
\begin{tikzpicture}[
    >=Stealth,
    node distance=0.4cm and 0.4cm,
    % === Unified Node Styles ===
    stdbox/.style={
      draw, thick, minimum width=2.4cm,
      fill=gray!8, draw=black!60, font=\scriptsize, align=center, rounded corners=3pt, inner sep=3pt},
    vlmbox/.style={
      draw, thick, minimum width=2.4cm,
      fill=orange!8, draw=orange!80!black, font=\scriptsize, align=center, rounded corners=3pt, inner sep=3pt},
    revbox/.style={
      draw, thick, minimum width=2.0cm, minimum height=1.0cm,
      fill=red!8, draw=red!80!black, font=\scriptsize, align=center, rounded corners=3pt, inner sep=3pt},
    distbox/.style={
      draw, thick, minimum width=2.0cm, minimum height=1.0cm,
      fill=green!10, draw=green!60!black, font=\scriptsize, align=center, rounded corners=3pt, inner sep=3pt},
    lbl/.style={font=\sffamily\bfseries\small, text=black!70},
    arr/.style={->, thick, black!70},
    condarr/.style={->, thick, dashed, red!60!black},
]

% =====================================================================
%  TOP ROW -- Main (fast) path
% =====================================================================

\node[stdbox] (wsi) {\includegraphics[width=2.2cm,height=1.45cm,keepaspectratio]{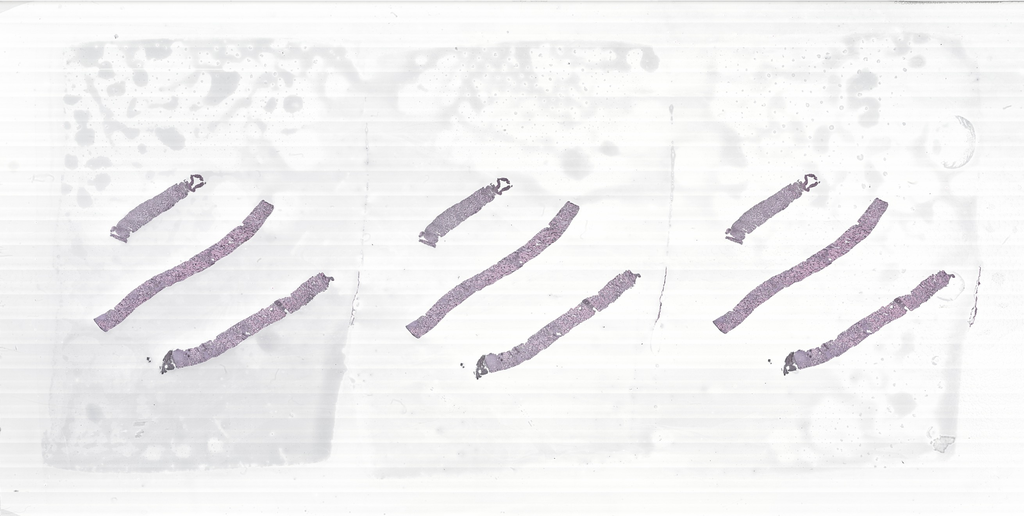}\\[0.5ex] \textbf{WSI Input}\\[-0.5ex] \textcolor{black!60}{\tiny thumbnail extraction}};

\node[vlmbox, right=of wsi] (bbox) {\includegraphics[width=2.2cm,height=1.45cm,keepaspectratio]{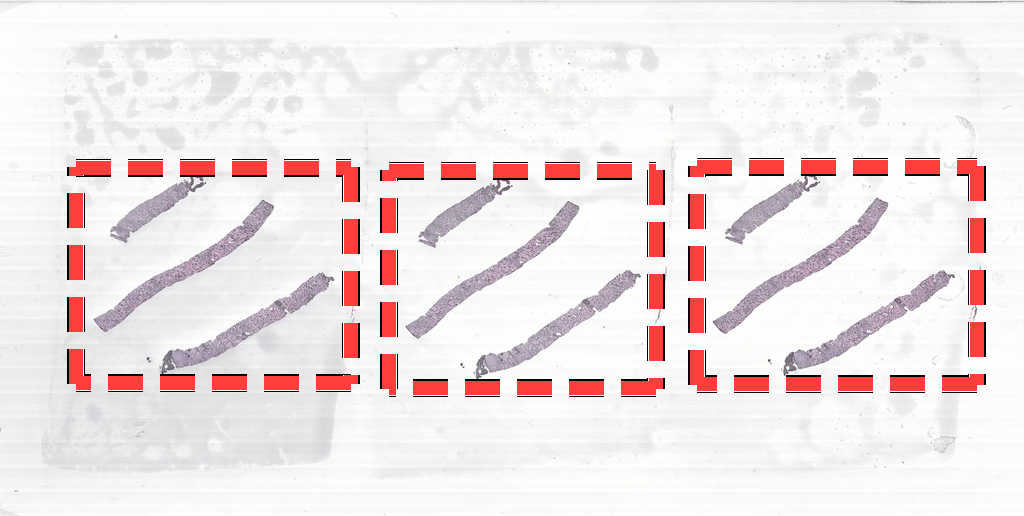}\\[0.5ex] \textbf{Box Grounding}\\[-0.5ex] \textcolor{orange!80!black}{\tiny visual grounding}};

\node[stdbox, right=of bbox] (crop) {\includegraphics[width=2.2cm,height=1.45cm,keepaspectratio]{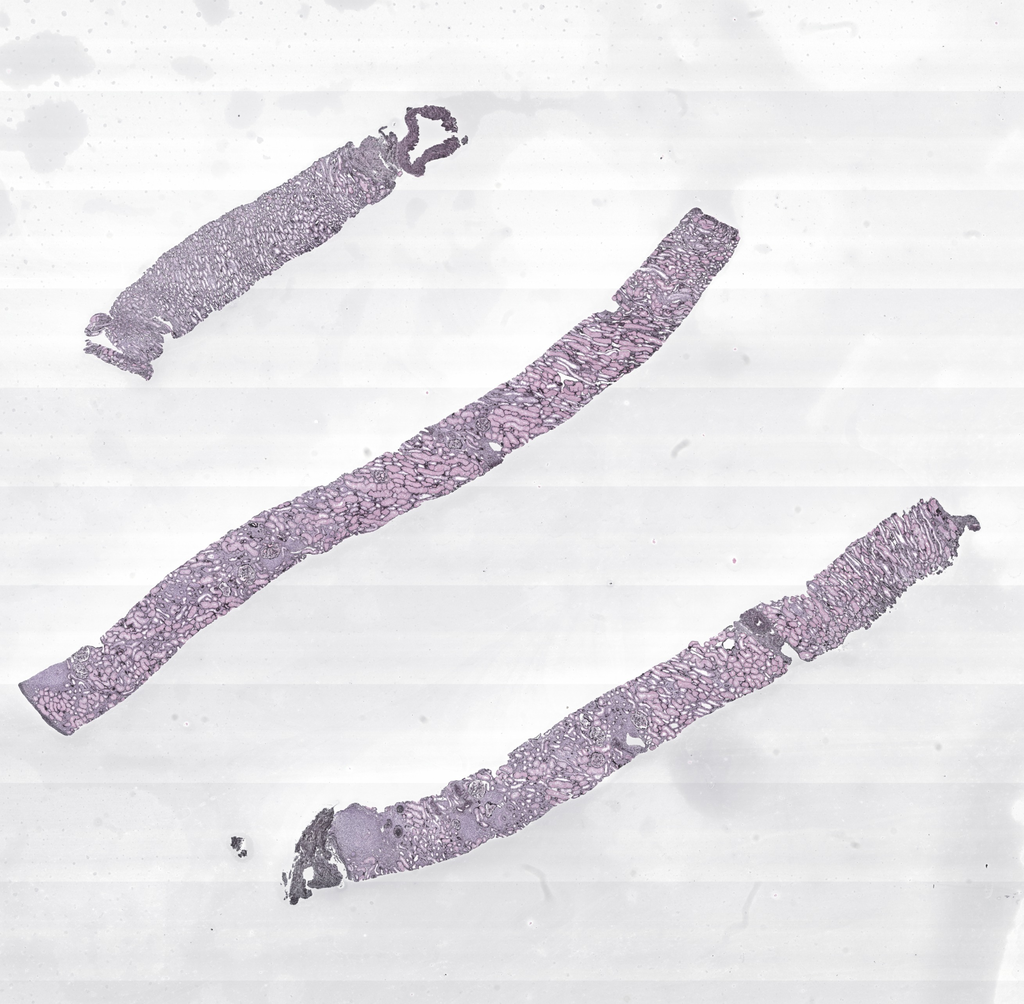}\\[0.5ex] \textbf{Tissue Crops}\\[-0.5ex] \textcolor{black!60}{\tiny local isolation}};

\node[stdbox, right=of crop] (fgmask) {\includegraphics[width=2.2cm,height=1.45cm,keepaspectratio]{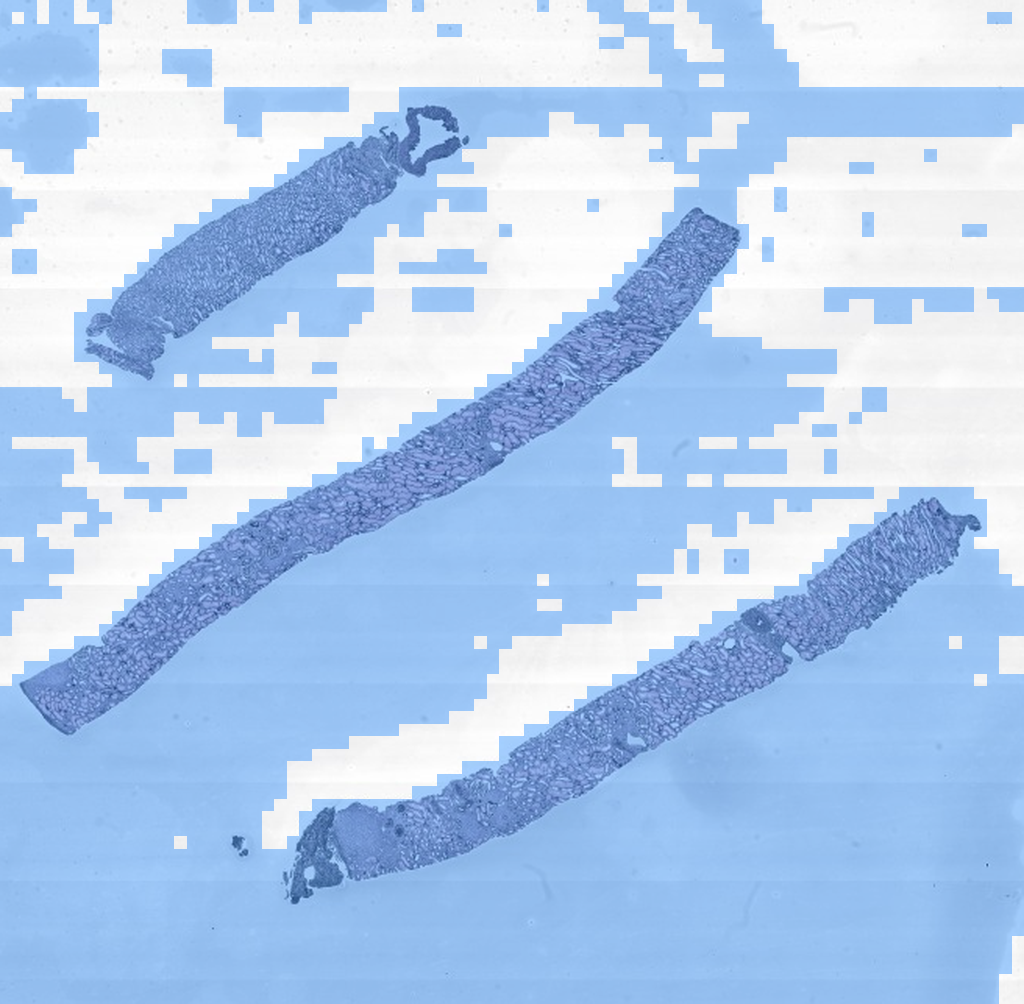}\\[0.5ex] \textbf{Colour Seg.}\\[-0.5ex] \textcolor{black!60}{\tiny two-pass $K$-means}};

\node[vlmbox, right=of fgmask] (zslabels) {\includegraphics[width=2.2cm,height=1.45cm,keepaspectratio]{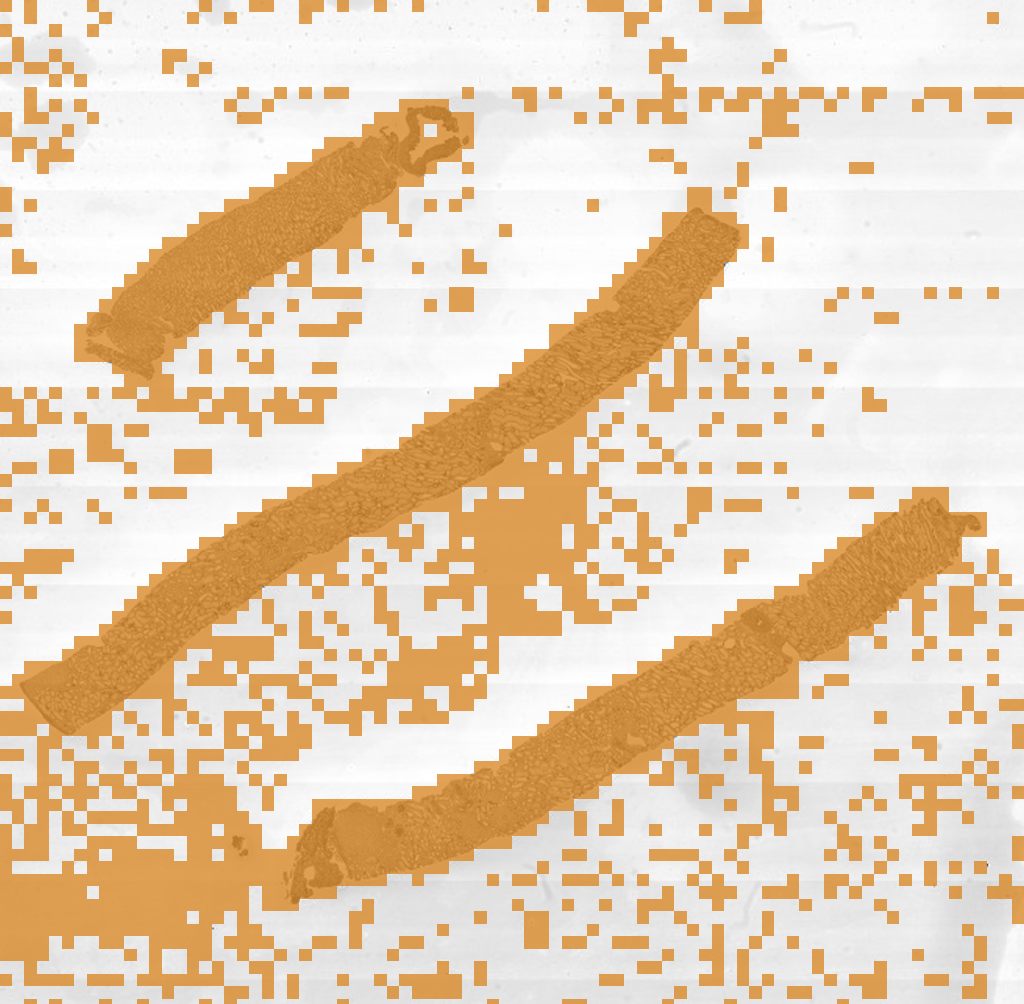}\\[0.5ex] \textbf{Zero-Shot VLM}\\[-0.5ex] \textcolor{orange!80!black}{\tiny patch classification}};

\node[stdbox, right=of zslabels] (pptop) {\includegraphics[width=2.2cm,height=1.45cm,keepaspectratio]{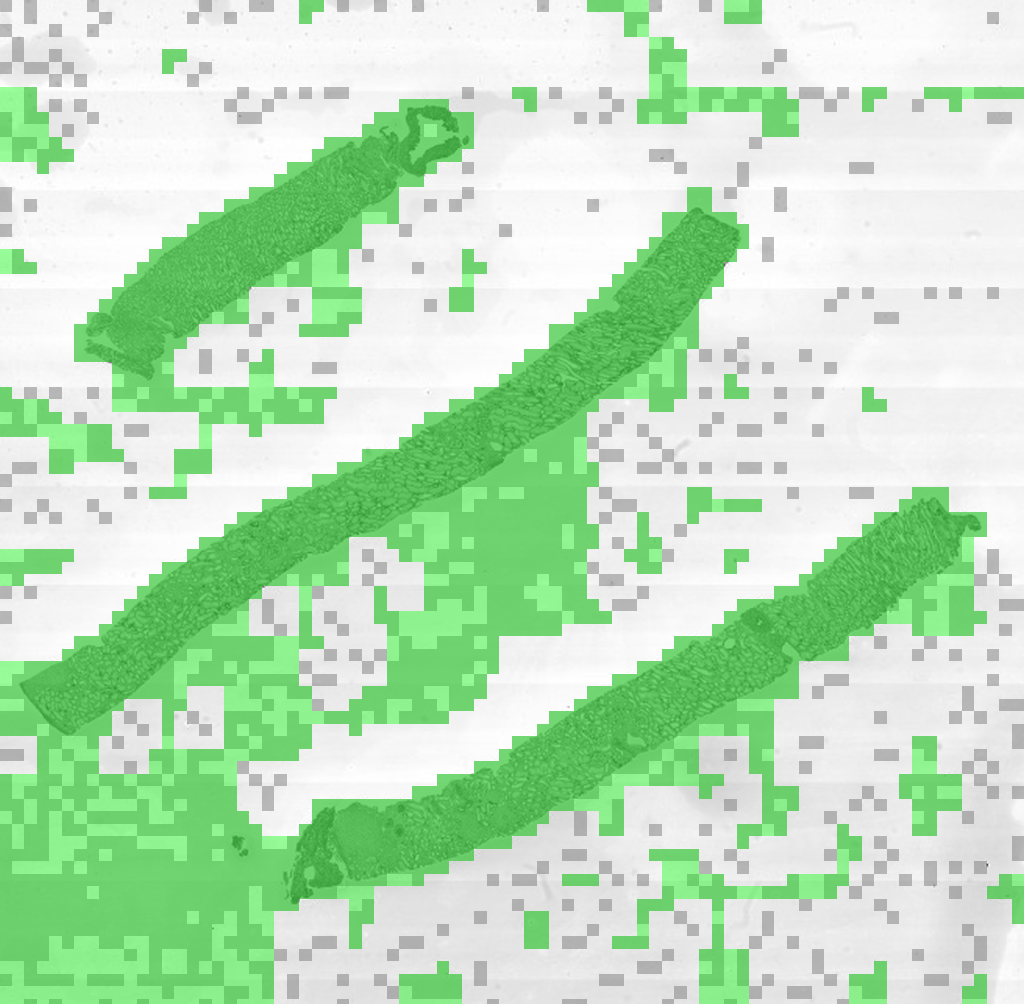}\\[0.5ex] \textbf{Refinement}\\[-0.5ex] \textcolor{black!60}{\tiny morphological ops.}};

% Top row connections
\draw[arr] (wsi) -- (bbox);
\draw[arr] (bbox) -- (crop);
\draw[arr] (crop) -- (fgmask);
\draw[arr] (fgmask) -- (zslabels);
\draw[arr] (zslabels) -- (pptop);

% =====================================================================
%  REVIEWER & DISTILLATION
% =====================================================================
\node[revbox, right=0.6cm of pptop] (reviewer) {\textbf{VLM Reviewer}\\[-0.2ex] \textcolor{red!80!black}{\tiny quality audit}};
\node[distbox, right=0.8cm of reviewer] (distill) {\textbf{Student}\\[-0.2ex]\textbf{Distillation}\\[-0.2ex] \textcolor{green!40!black}{\tiny lightweight model}};

\draw[arr] (pptop) -- node[midway, above, font=\bfseries\footnotesize, text=red!70!black] {\texttimes} (reviewer);
\draw[arr] (reviewer) -- node[midway, above, font=\tiny\sffamily\bfseries, text=green!55!black] {pass} (distill);

% =====================================================================
%  BOTTOM ROW -- Auto Context 
% =====================================================================
% MOVED DOWN: Changed from below=1.0cm to below=1.2cm
\node[vlmbox, below=1.2cm of crop] (ptoverlay) {\includegraphics[width=2.2cm,height=1.45cm,keepaspectratio]{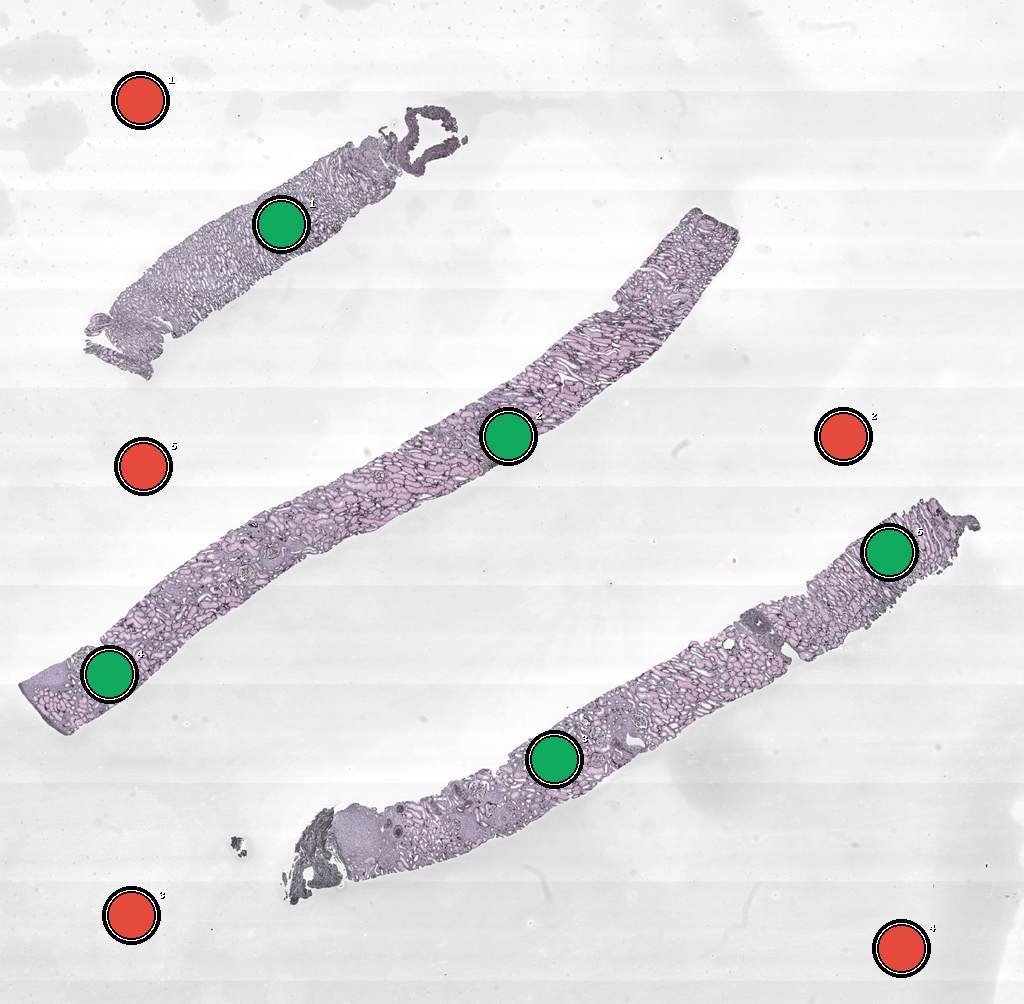}\\[0.5ex] \textbf{Point Grounding}\\[-0.5ex] \textcolor{orange!80!black}{\tiny candidate sampling}};

\node[vlmbox, right=of ptoverlay] (exemplars) {\includegraphics[width=2.2cm,height=1.45cm,keepaspectratio]{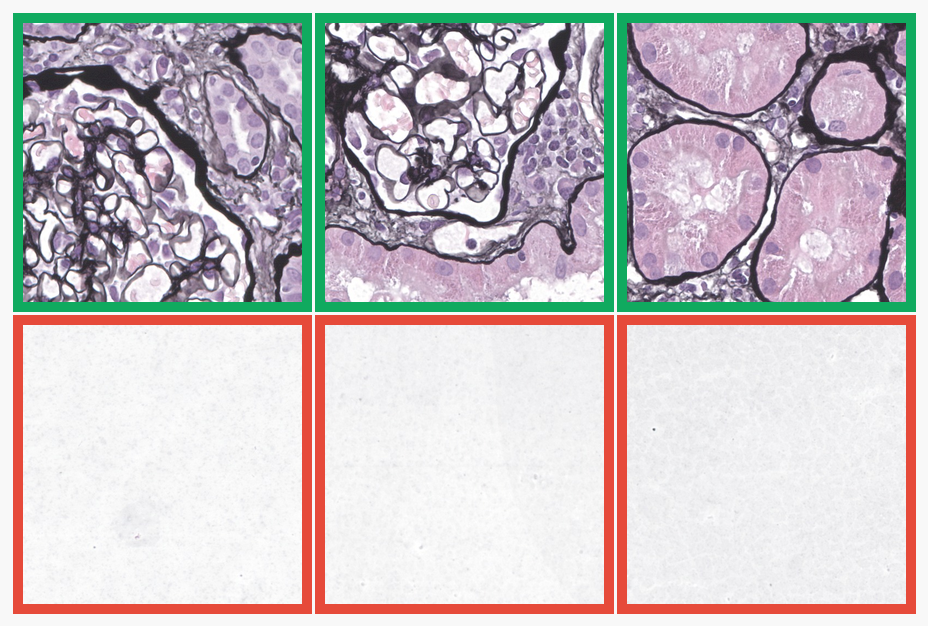}\\[0.5ex] \textbf{Top-$K$ Selection}\\[-0.5ex] \textcolor{orange!80!black}{\tiny exemplar ranking}};

\node[vlmbox, right=of exemplars] (fslabels) {\includegraphics[width=2.2cm,height=1.45cm,keepaspectratio]{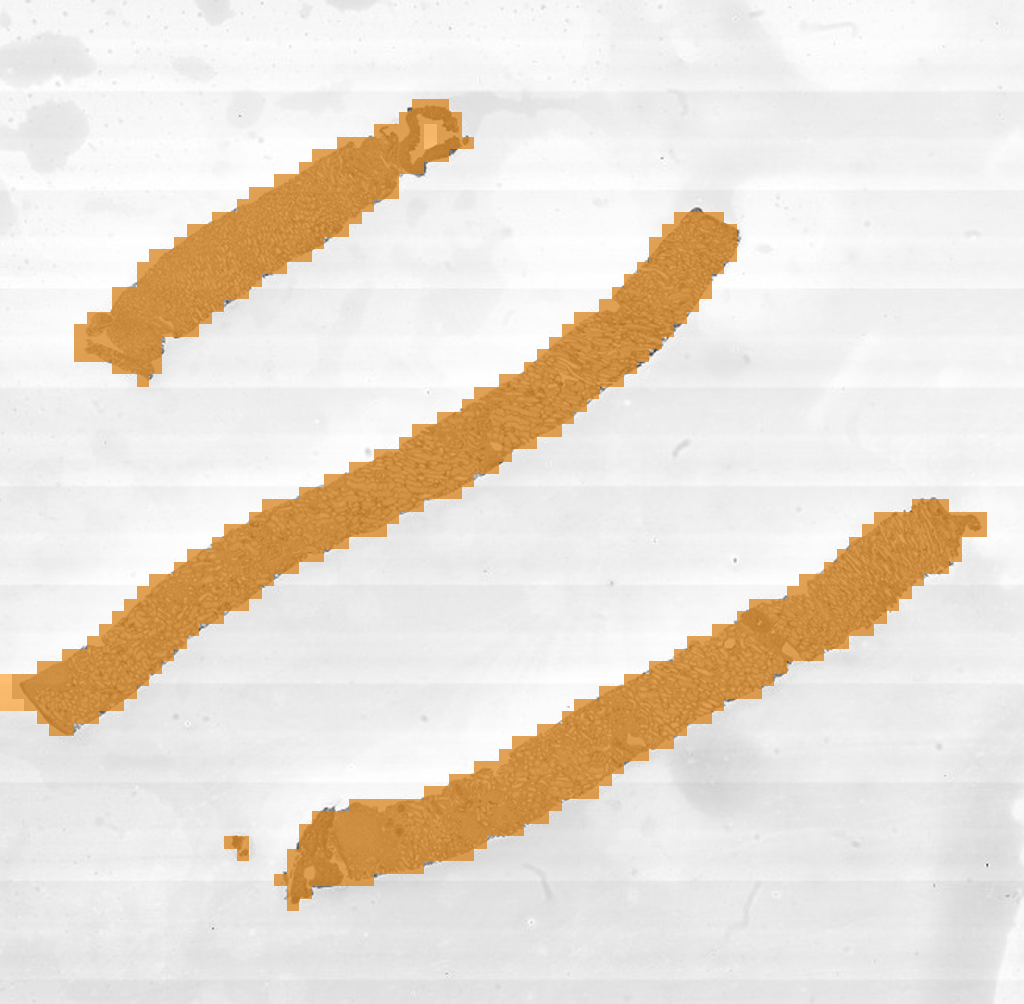}\\[0.5ex] \textbf{Few-Shot VLM}\\[-0.5ex] \textcolor{orange!80!black}{\tiny in-context inference}};

\node[stdbox, right=of fslabels] (ppbot) {\includegraphics[width=2.2cm,height=1.45cm,keepaspectratio]{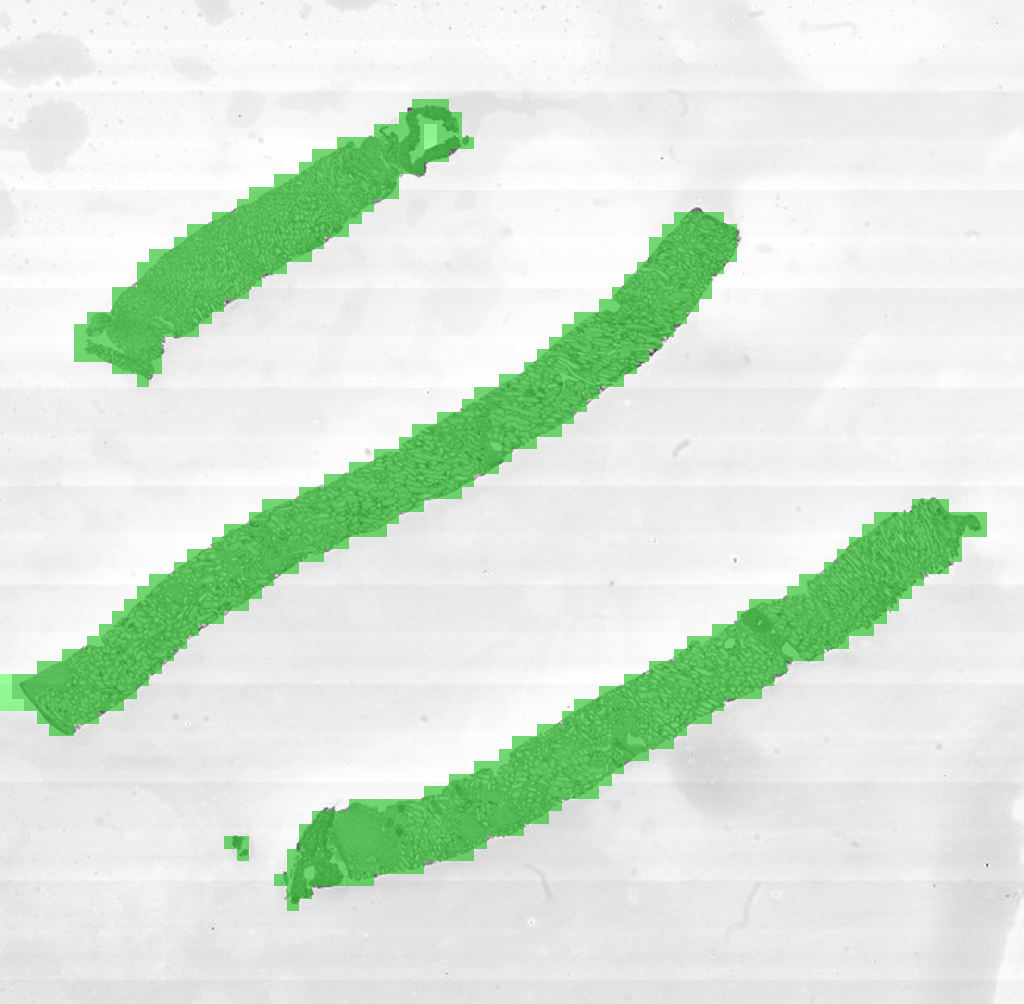}\\[0.5ex] \textbf{Refinement}\\[-0.5ex] \textcolor{black!60}{\tiny morphological ops.}};

% Bottom row connections
\draw[arr] (ptoverlay) -- (exemplars);
\draw[arr] (exemplars) -- (fslabels);
\draw[arr] (fslabels) -- (ppbot);

% Connection from Reviewer Fail to Auto Context
% ADJUSTED: Changed vertical drop from -0.65 to -0.85 so it stays centered
\draw[condarr] (reviewer.south) -- ++(0,-0.85) -| (ptoverlay.north)
    node[pos=0.15, below right, font=\tiny\sffamily\bfseries] {fail (low quality)};

% Auto Context Pass -> Distill 
% FIXED: Cleaned up path to ensure it goes right (-), then perfectly UP (|) into the bottom center.
\draw[arr] (ppbot.east) -| node[pos=0.25, above, font=\bfseries\footnotesize, text=green!55!black] {\checkmark} (distill.south);

% --- Highlight Box ---
\begin{scope}[on background layer]
   \node[draw=orange!70!black, dashed, thick, rounded corners=6pt, 
         fill=orange!4, inner sep=12pt, 
         fit=(ptoverlay)(ppbot), 
         label={[font=\sffamily\bfseries, text=orange!70!black, anchor=south west, inner sep=2pt]north west:Auto Context (Few-Shot Refinement)}]
     (acbox) {};
\end{scope}

\end{tikzpicture}
}
\caption{ 
The primary fast path (top) uses VLM-guided region proposals and zero-shot patch classification to generate initial tissue masks. A VLM reviewer (red) automatically audits quality; failed masks are routed to an \textit{Auto Context} loop (bottom) where slide-specific exemplars are mined to rescue performance via few-shot refinement. Validated pseudo-labels are ultimately distilled into a lightweight student model for deployment.}
\label{fig:pipeline}
\end{figure}
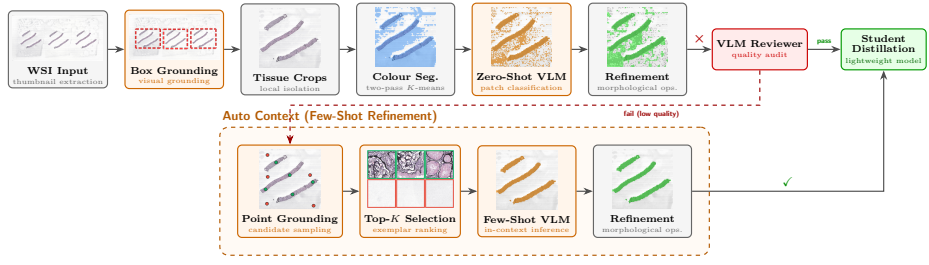

%% file: evaluation.tex
\section{Evaluation}
\label{sec:evaluation}

%\noindent\textbf{Experimental setup.}
We evaluate on \textit{Leica-75}: 75 renal transplant WSIs (25 per stain: EVG, H\&E, Jones) scanned on a Leica system. Slides are partitioned into non-overlapping $512{\times}512$ level-0 patches and evaluated against manually curated foreground masks spanning 220 annotated tissue cores. We additionally define a curated stress-test subset, \textit{Stress-32} ($n{=}32$; 23 Jones, 9 EVG), comprising slides where background artefacts can be confused with tissue under high magnification. For blur assessment, we sample 200 patches from 186 WSIs, stratified by Qwen3 VL 8B predicted focus labels (sharp / mild blur / out of focus) into approximately equal thirds, and collect independent human judgements for each patch. For evaluating VLMs as segmentation-mask reviewers, we discretise precision and recall into four grades (excellent $\geq$95\%, good 80--95\%, partial 50--80\%, poor $<$50\%) and construct a balanced benchmark by manually corrupting expert-verified masks to obtain uniform coverage over the resulting $4{\times}4{\times}3$ grid (precision $\times$ recall $\times$ stain), yielding 96 masks with known ground-truth grades. We distil two student models using Leica-75 or Stress-32 following a 70/10/20 train/val/test split, sampling 5{,}000 foreground and background patches from the training split with standard data augmentations (colour jitter, random rotations, flips). We compare five baselines (Otsu~\cite{otsu1979threshold}, $K$-means~\cite{MacQueen1967SomeMF}, HistoQC~\cite{janowczyk2019histoqc}, TRIDENT-GrandQC~\cite{weng2024grandqc,zhang2025accelerating}, TRIDENT-HEST~\cite{jaume2024hest,zhang2025accelerating}) against four VLM oracle configurations (Qwen3 VL 2B/8B, zero-shot and few-shot \cite{bai2025qwen3vltechnicalreport}). For few-shot Auto-context, we set $K{=}1$. Statistical comparisons use paired Wilcoxon signed-rank tests with Bonferroni correction. Finally, we evaluate 264 WSIs from an external kidney-pathology centre in a different country.

%% =========================================================================
%%  Tab. 1: Main Dice (full width, headline result)
%% =========================================================================
\begin{table}[t]
\centering
\caption{Dice (mean$_{\pm\text{std}}$) on 75 Leica WSIs. Best per stain \textbf{bold}. $^\dagger$Paired Wilcoxon $p{<}0.01$ vs.\ best non-VLM baseline ($p{<}0.001$ EVG; $p{=}0.019$ H\&E; $p{=}0.542$ Jones).}
\label{tab:dice-main}
\scriptsize
\setlength{\tabcolsep}{2pt}
\resizebox{\textwidth}{!}{%
\begin{tabular}{l ccccc cccc}
\toprule
& \multicolumn{5}{c}{\textit{Baselines}} & \multicolumn{4}{c}{\textit{Ours (VLM oracle)}} \\
\cmidrule(lr){2-6}\cmidrule(lr){7-10}
Stain & Otsu & $K$-means & HistoQC & GrandQC & HEST & 2B-ZS & 8B-ZS & 2B-FS & 8B-FS \\
\midrule
EVG   & $0.638_{\pm.284}$ & $0.667_{\pm.270}$ & $0.548_{\pm.276}$ & $0.769_{\pm.148}$ & $0.793_{\pm.049}$ & $0.835_{\pm.123}$ & $\mathbf{0.853_{\pm.041}}^{\dagger}$ & $0.844_{\pm.039}$ & $0.847_{\pm.035}$ \\
H\&E  & $0.679_{\pm.204}$ & $0.667_{\pm.224}$ & $0.224_{\pm.235}$ & $\mathbf{0.870_{\pm.047}}$ & $0.807_{\pm.075}$ & $0.852_{\pm.034}$ & $0.843_{\pm.045}$ & $0.829_{\pm.047}$ & $0.828_{\pm.047}$ \\
Jones & $0.644_{\pm.239}$ & $0.637_{\pm.244}$ & $0.281_{\pm.298}$ & $0.775_{\pm.194}$ & $0.743_{\pm.109}$ & $0.854_{\pm.072}$ & $\mathbf{0.858_{\pm.027}}$ & $0.843_{\pm.031}$ & $0.840_{\pm.031}$ \\
\bottomrule
\end{tabular}%
}
\end{table}

\noindent\textbf{Foreground segmentation.}
Tab.~\ref{tab:dice-main} reports stain-stratified Dice. On both specialized stains, VLM methods achieve the highest performance: Qwen\,8B-ZS reaches $0.853_{\pm0.041}$ on EVG and $0.858_{\pm0.027}$ on Jones, significantly outperforming the best non-VLM baselines (HEST $0.793$ on EVG, $p{<}0.001$; GrandQC $0.775$ on Jones). On in-distribution H\&E, GrandQC achieves $0.870_{\pm0.047}$ owing to its supervised training advantage; VLM methods remain competitive at $0.852_{\pm0.034}$ without any in-context examples.
HistoQC achieves moderate precision but catastrophically low recall ($0.150$ on H\&E, $0.209$ on Jones), discarding the majority of viable tissue.
Aggregating across stains, our 8B-ZS oracle achieves the highest mean Dice ($\bar{D}{=}0.851$) with $7{\times}$ lower cross-stain standard deviation ($\sigma_s{=}0.006$) than GrandQC ($\sigma_s{=}0.046$, $D_{\min}{=}0.769$), confirming more consistent predictions across stain families. To verify that this is not an artefact of the stages preceding the VLM, we ran a fairness ablation giving GrandQC identical bounding-box and colour-mask preprocessing: Qwen-8B zero-shot still yields higher recall on all three stains ($\Delta R{=}+0.115$ EVG, $+0.040$ H\&E, $+0.133$ Jones), confirming the gain originates from the VLM decision layer itself. Under-segmentation is the more dangerous failure mode because it irrecoverably excludes diagnostically relevant patches from all downstream analysis.

%% =========================================================================
%%  Tab. 2: minipage -- (a) Component ablation | (b) Stress-32
%% =========================================================================
\begin{table}[t]
\begin{minipage}[t]{0.48\textwidth}
\centering
\caption{Ablation on Stress-32 test ($n{=}32$). Each row adds one stage.}
\label{tab:ablation}
\scriptsize
\setlength{\tabcolsep}{1.5pt}
\resizebox{\textwidth}{!}{%
\begin{tabular}{cl cccc c}
  \toprule
  \# & Configuration & Dice & Prec & Rec & Acc & $\Delta$D \\
  \midrule
  \rowcolor{gray!5}
  1 & 2B WSI-level ZS      & $.107_{\pm.060}$ & $.058_{\pm.034}$ & $.976_{\pm.019}$ & $.735_{\pm.073}$ & {--} \\
  2 & $+$\,VLM bbox filter & $.426_{\pm.107}$ & $.279_{\pm.092}$ & $.976_{\pm.019}$ & $.958_{\pm.028}$ & $+.319$ \\
  \rowcolor{gray!5}
  3 & $+$\,Colour filter   & $.520_{\pm.159}$ & $.378_{\pm.176}$ & $.952_{\pm.034}$ & $.971_{\pm.020}$ & $+.094$ \\
  4 & $+$\,Auto-context 2B & $.812_{\pm.093}$ & $.811_{\pm.133}$ & $.832_{\pm.093}$ & $.994_{\pm.005}$ & $+.292$ \\
  \rowcolor{gray!5}
  5 & $+$\,Auto-context 8B & $.836_{\pm.063}$ & $.829_{\pm.073}$ & $.846_{\pm.076}$ & $.995_{\pm.004}$ & $+.024$ \\
  6 & $+$\,Post-processing & $\mathbf{.846_{\pm.057}}$ & $.832_{\pm.067}$ & $.863_{\pm.069}$ & $.995_{\pm.004}$ & $
  +.010$ \\
  \bottomrule
  \end{tabular}%
}
\end{minipage}%
\hfill
\begin{minipage}[t]{0.50\textwidth}
\centering
\caption{Stress-32 ($n{=}32$): zero-shot vs.\ few-shot. $^{***}p{<}0.001$ (paired Wilcoxon).}
\label{tab:harder-jones}
\scriptsize
\vspace{-0.03cm}
\setlength{\tabcolsep}{1.5pt}
\resizebox{\textwidth}{!}{%
\begin{tabular}{l cccc cc}
\toprule
& \multicolumn{4}{c}{\textit{Absolute}} & \multicolumn{2}{c}{\textit{$\Delta$(FS$-$ZS)}} \\
\cmidrule(lr){2-5}\cmidrule(lr){6-7}
 & 2B-ZS & 2B-FS & 8B-ZS & 8B-FS & $\Delta$2B & $\Delta$8B \\
\midrule
Dice & $.470_{\pm.172}$ & $.819_{\pm.094}$ & $.768_{\pm.107}$ & $\mathbf{.846_{\pm.057}}$ & $+.349^{***}$ & $+.078^{***}$ \\
Prec & $.337_{\pm.181}$ & $.810_{\pm.133}$ & $.691_{\pm.147}$ & $\mathbf{.832_{\pm.067}}$ & $+.473^{***}$ & $+.141^{***}$ \\
Rec  & $\mathbf{.949_{\pm.053}}$ & $.848_{\pm.089}$ & $.894_{\pm.064}$ & $.863_{\pm.069}$ & $-.101^{***}$ & $-.031^{***}$ \\
Acc  & $.972_{\pm.017}$ & $.995_{\pm.004}$ & $.993_{\pm.006}$ & $\mathbf{.996_{\pm.002}}$ & $+.023^{***}$ & $+.003^{***}$ \\
\bottomrule
\end{tabular}%
}
\end{minipage}
\end{table}

\noindent\textbf{Component ablation.}
Tab.~\ref{tab:ablation} quantifies the contribution of each pipeline stage on \textit{Stress-32}. Na\"ive WSI-level zero-shot classification (row~1) achieves $0.107_{\pm.060}$ Dice due to extreme base-rate imbalance: tissue typically occupies only a small fraction of the slide, so even a low false-positive rate yields many spurious foreground patches and collapses precision. VLM bounding-box filtering (row~2) provides the first major gain ($\Delta D{=}+0.319$) by restricting processing to VLM-localised tissue regions. Two-pass colour refinement adds a further $+0.094$ (row~3). The largest improvement comes from \textit{Auto-context} exemplar mining (K=1) and few-shot prompting (row~4), which increases Dice by $+0.292$ and improves precision from $0.378$ to $0.811$ while maintaining high recall ($0.832$). Scaling from 2B to 8B yields an additional $+0.024$ (row~5), and lightweight morphological post-processing contributes a final $+0.010$ to reach $0.846_{\pm.057}$ Dice (row~6).

\noindent\textbf{Few-shot stress test.}
On the stress-32 set (Tab.~\ref{tab:harder-jones}), Qwen\,2B-ZS achieves $0.470_{\pm0.172}$ Dice (precision $0.337$) due to deceptive background textures, while Qwen\,8B-ZS reaches $0.768_{\pm0.107}$. Few-shot prompting with automatically curated exemplars yields striking improvements: Qwen\,2B rises to $0.819_{\pm0.094}$ ($\Delta{=}+0.349$, $p{<}0.001$) and Qwen\,8B to $0.846_{\pm0.057}$ ($\Delta{=}+0.078$, $p{<}0.001$). The mechanism is precision-driven ($+0.473$ for 2B): the model learns to reject deceptive background patches by leveraging slide-specific exemplars, with individual case gains exceeding $+0.6$ Dice. %The slight recall decrease ($-0.101$) is a favourable trade-off confirmed by the large net Dice improvement.

\input{figures/qualitative}

%% =========================================================================
%%  Tab. 3: minipage -- (a) Distillation | (b) Blur
%% =========================================================================
\begin{table}[t]
\begin{minipage}[t]{0.55\textwidth}
\centering
\caption{Knowledge distillation: MobileNetV3 student (4.2\,M) vs.\ VLM teacher ($\sim$8B). $^\dagger$Patches/sec, single GPU.}
\label{tab:distill}
\scriptsize
\setlength{\tabcolsep}{1.5pt}
\resizebox{\textwidth}{!}{%
\begin{tabular}{ll ccccc c}
\toprule
Eval & Model & Params & $F1\!\uparrow$ & IoU$\!\uparrow$ & $P\!\uparrow$ & $R\!\uparrow$ & p/s$^\dagger$ \\
\midrule
\rowcolor{gray!5}
\multirow{2}{*}{\scriptsize Leica-75} & Qwen-8B ZS teach. & $\sim$8B & .877 & .781 & .942 & .821 & 6 \\
& Student (Leica-75) & 4.2M & $\mathbf{.884}$ & $\mathbf{.792}$ & .932 & .840 & 586 \\
\midrule
\rowcolor{gray!5}
\multirow{2}{*}{\scriptsize Stress-32} & Qwen-8B FS teach. & $\sim$8B & .894 & .809 & .926 & .864 & 4 \\
& Student (Stress-32) & 4.2M & $\mathbf{.907}$ & $\mathbf{.829}$ & .888 & .927 & 586 \\
\midrule
\rowcolor{gray!5}
\multicolumn{2}{l}{\scriptsize Student (Stress-32)$\to$Leica-75 test} & 4.2M & .881 & .787 & .929 & .837 & 586 \\
\multicolumn{2}{l}{\scriptsize Student (Leica-75)$\to$Stress-32 test} & 4.2M & .701 & .539 & .579 & .887 & 586 \\
\bottomrule
\end{tabular}%
}
\end{minipage}%
\hfill
\begin{minipage}[t]{0.43\textwidth}
\centering
\caption{Blur detection ($n{=}200$): Qwen\,8B vs.\ human consensus.}
\label{tab:blur}
\scriptsize
\setlength{\tabcolsep}{2pt}
\vspace{0.37cm}
\resizebox{\textwidth}{!}{%
\begin{tabular}{lrc}
\toprule
Metric & Est. & 95\,\% CI \\
\midrule
Hum.\ $\alpha$ (3-cl.) & $.927_{\pm.016}$ & $[.894,.955]$ \\
VLM $\kappa_w$ (3-cl.)  & $.708_{\pm.036}$ & $[.632,.773]$ \\
VLM $\kappa$ (bin.)      & $.989_{\pm.009}$ & $[.963,1.00]$ \\
VLM sens.\ (bin.)        & $1.00_{\pm.000}$ & $[1.00,1.00]$ \\
VLM spec.\ (bin.)        & $.993_{\pm.006}$ & $[.977,1.00]$ \\
\bottomrule
\end{tabular}
}%
\end{minipage}
\end{table}

\noindent\textbf{Knowledge distillation.}
Tab.~\ref{tab:distill} shows that a lightweight MobileNetV3 student (4.2\,M parameters) trained purely on VLM pseudo-labels matches or exceeds its Qwen-8B teacher on held-out data, while running ${\sim}100{\times}$ faster (586 vs.\ 4--6 patches/sec). On the Leica benchmark the student achieves $F1{=}0.884$ (teacher $0.877$); on the stress set $F1{=}0.907$ (teacher $0.894$). Cross-set transfer (bottom rows) reveals that a student distilled on stress-32 generalises well to Leica-75 ($F1{=}0.881$), but the reverse degrades ($F1{=}0.701$), confirming that per-site distillation on representative data is important for robust deployment.

\noindent\textbf{Blur detection.}
For focus quality assessment (Tab.~\ref{tab:blur}), Qwen\,8B achieves near-perfect binary agreement with human consensus ($\kappa{=}0.989$, sensitivity $1.000$, specificity $0.993$) and strong 3-class ordinal agreement ($\kappa_w{=}0.708$) with zero extreme misclassifications. This confirms that VLMs can serve as reliable automated quality gates for out-of-focus patches at no additional inference cost, since blur labels are extracted jointly with tissue classifications.

%% =========================================================================
%%  Tab. 4: Reviewer alignment (full width, individual + group means)
%% =========================================================================
\begin{table}[t]
\centering
\caption{Mask-quality reviewer alignment on  corrupted masks ($n{=}96$). Best VLM \textbf{bold}; best human \underline{underlined}; $^{\ddagger}$VLM exceeds best human. Bottom rows: group means.}
\label{tab:reviewer}
\scriptsize
\setlength{\tabcolsep}{2pt}
\resizebox{\textwidth}{!}{%
\begin{tabular}{l ccc cc cc}
\toprule
& \multicolumn{3}{c}{\textit{Accuracy (grade-level)}} & \multicolumn{2}{c}{\textit{QWK (ordinal)}} & \multicolumn{2}{c}{\textit{MAE\,(\%)\,$\downarrow$}} \\
\cmidrule(lr){2-4}\cmidrule(lr){5-6}\cmidrule(lr){7-8}
Rater & Prec & Rec & Joint & Prec & Rec & Prec & Rec \\
\midrule
Hum.\,(Gi) & $\underline{.760_{\pm.089}}$ & $.750_{\pm.089}$ & $.604_{\pm.094}$ & $.886_{\pm.051}$ & $.896_{\pm.040}$ & $\underline{4.35_{\pm0.96}}$ & $4.35_{\pm1.03}$ \\
Hum.\,(C)  & $.729_{\pm.089}$ & $\underline{.854_{\pm.073}}$ & $\underline{.646_{\pm.099}}$ & $\underline{.887_{\pm.045}}$ & $\underline{.941_{\pm.034}}$ & $6.30_{\pm0.96}$ & $3.97_{\pm0.92}$ \\
Hum.\,(V)  & $.688_{\pm.099}$ & $.844_{\pm.073}$ & $.594_{\pm.099}$ & $.862_{\pm.051}$ & $.935_{\pm.032}$ & $6.09_{\pm1.43}$ & $\underline{3.26_{\pm0.65}}$ \\
\midrule
Gem.\,3.1\,Pro & $.740_{\pm.089}$ & $\mathbf{.885_{\pm.062}}^{\ddagger}$ & $.635_{\pm.094}$ & $.892_{\pm.045}^{\ddagger}$ & $\mathbf{.953_{\pm.030}}^{\ddagger}$ & $7.55_{\pm1.39}$ & $4.58_{\pm0.85}$ \\
Gem.\,3\,Flash & $.781_{\pm.083}^{\ddagger}$ & $.875_{\pm.068}^{\ddagger}$ & $.667_{\pm.089}^{\ddagger}$ & $\mathbf{.896_{\pm.053}}^{\ddagger}$ & $.951_{\pm.031}^{\ddagger}$ & $\mathbf{6.09_{\pm1.11}}$ & $\mathbf{3.99_{\pm0.62}}$ \\
Gem.\,3\,Pro   & $\mathbf{.792_{\pm.083}}^{\ddagger}$ & $.875_{\pm.068}^{\ddagger}$ & $\mathbf{.708_{\pm.094}}^{\ddagger}$ & $.894_{\pm.054}^{\ddagger}$ & $.915_{\pm.080}$ & $7.49_{\pm1.28}$ & $4.82_{\pm1.25}$ \\
\midrule
\rowcolor{gray!5}
\textit{Human mean} & \textit{.726} & \textit{.816} & {--} & \textit{.878} & \textit{.924} & \textit{5.58} & \textit{3.86} \\
\rowcolor{gray!5}
\textit{VLM mean}   & \textit{.771} & \textit{.878} & {--} & \textit{.894} & \textit{.939} & \textit{7.04} & \textit{4.46} \\
\bottomrule
\end{tabular}%
}
\end{table}

\noindent\textbf{VLM reviewer alignment.}
For mask-quality grading (Tab.~\ref{tab:reviewer}), Gemini\,3\,Pro achieves the highest joint exact-match accuracy ($0.708_{\pm0.094}$), exceeding the best human rater ($0.646_{\pm0.099}$) by $+0.062$. Group means show VLMs outperform humans on grade-level accuracy ($0.771/0.878$ vs.\ $0.726/0.816$ for precision/recall) and ordinal agreement (QWK $0.894/0.939$ vs.\ $0.878/0.924$), while humans retain a calibration advantage (MAE $5.58$ vs.\ $7.04$). Inter-rater QWK averages $0.904$ for human-human, $0.884$ for VLM-VLM, and $0.857$ for human-VLM pairs. %, confirming VLM reviewers operate within the range of expert variability. 
On a complementary subjective review task, collapsing grades to clinically actionable bins (excellent vs.\ needs review), VLM-human agreement remains substantial ($\kappa{=}0.717$), confirming VLMs can reliably triage masks for human escalation even when given no explicit scoring criteria.% to evaluate quality of segmentation masks.

\noindent\textbf{Multi-centre validation.}
We ran our implementation at an external center abroad, on a cohort from a separate kidney-pathology center (264 WSIs, 353 valid bounding boxes across six hospital sites). Qwen-8B zero-shot achieves an overall Dice of $0.784_{\pm0.199}$ with high recall ($0.964_{\pm0.096}$) and moderate precision ($0.701_{\pm0.205}$). Per-site Dice ranges from $0.715$ to $0.840$. The lower precision relative to the Leica-75 benchmark is expected, as this external cohort was curated as a deliberately challenging evaluation set. The consistently high recall ($>0.94$ at all sites) confirms that our method reliably detects tissue across scanners and institutions without site-specific tuning.

\noindent\textbf{Discussion.}
General-purpose VLMs, despite no pathology-specific training, surpass dedicated computational pathology tools on out-of-distribution stains with $7{\times}$ lower cross-stain variance, validating the thesis that foreground segmentation is a perceptual task. Few-shot prompting with automatically curated exemplars addresses the long tail of difficult cases through per-slide adaptation at inference time, without retraining or human intervention. VLM-based annotation review achieves agreement with ground truth that is statistically indistinguishable from human expert consensus, and distilled students preserve teacher-level performance at ${\sim}100{\times}$ the throughput.

%% file: figures/qualitative.tex
% Qualitative figure intentionally omitted from this arXiv source bundle.